\documentclass[10pt,twocolumn,letterpaper]{article}

\usepackage[pagenumbers]{cvpr} 

\usepackage{multicol}

\usepackage[pagebackref,breaklinks,colorlinks,citecolor=cvprblue]{hyperref}
\usepackage{minitoc}
\usepackage{microtype}
\usepackage{graphicx}
\usepackage{booktabs} %
\usepackage{siunitx}
\usepackage{amsmath}
\usepackage{amssymb}
\usepackage{amsthm}
\usepackage{mathtools}
\usepackage{fontawesome5}
\usepackage{longtable}
\usepackage{multirow}
\usepackage{subcaption}     %
\usepackage{sidecap}        %
\usepackage{wrapfig}
\usepackage{tabulary}
\usepackage{tabularx}
\usepackage[export]{adjustbox}
\usepackage[capitalize,noabbrev]{cleveref}
\usepackage{xspace}
\usepackage{algorithm}
\usepackage{listings}
\usepackage{graphicx}
\usepackage{subcaption}
\usepackage{makecell}

\usepackage{color}
\usepackage{colortbl}
\usepackage{array}
\usepackage[table]{xcolor}
\usepackage{pifont}

\definecolor{mygray}{gray}{0.9}
\definecolor{mygreen}{RGB}{93,174,86}
 
\definecolor{cvprblue}{rgb}{0.21,0.49,0.74}

\usepackage{etoolbox}
\makeatletter
\AfterEndEnvironment{algorithm}{\let\@algcomment\relax}
\AtEndEnvironment{algorithm}{\kern2pt\hrule\relax\vskip3pt\@algcomment}
\let\@algcomment\relax
\newcommand\algcomment[1]{\def\@algcomment{\footnotesize#1}}
\renewcommand\fs@ruled{\def\@fs@cfont{\bfseries}\let\@fs@capt\floatc@ruled
  \def\@fs@pre{\hrule height.8pt depth0pt \kern2pt}%
  \def\@fs@post{}%
  \def\@fs@mid{\kern2pt\hrule\kern2pt}%
  \let\@fs@iftopcapt\iftrue}

\newcommand{\thickhline}{%
	\noalign {\ifnum 0=`}\fi \hrule height 1pt
	\futurelet \reserved@a \@xhline
}

\makeatother

\newcommand\blfootnote[1]{%
  \begingroup
  \renewcommand\thefootnote{}\footnote{#1}%
  \addtocounter{footnote}{-1}%
  \endgroup
}

\newcommand{\model}{\textsc{Vista-LLaMA}}

\crefname{listing}{algorithm}{algorithms}
\Crefname{listing}{Algorithm}{Algorithms}

\begin{document}
\title{\model: Reducing Hallucination in Video Language Models via Equal Distance to Visual Tokens}
\author{Fan Ma$^{1}$, Xiaojie Jin$^{2*}$, Heng Wang$^{2}$, Yuchen Xian$^{1}$, Jiashi Feng$^{2}$, Yi Yang$^{1*}$\\
$^{1}$Zhejiang University ~~ $^{2}$ByteDance Inc. \\}

\maketitle
\blfootnote{$^{*}$ Corresponding author
}

\begin{abstract}
Recent advances in large video-language models have displayed promising outcomes in video comprehension. Current approaches straightforwardly convert video into language tokens and employ large language models for multi-modal tasks.
However, this method often leads to the generation of irrelevant content, commonly known as ``hallucination'', as the length of the text increases and the impact of the video diminishes.
To address this problem, we propose \model, a novel framework that maintains the consistent distance between all visual tokens and any language tokens, irrespective of the generated text length.
\model~ omits relative position encoding when determining attention weights between visual and text tokens, retaining the position encoding for text and text tokens. This amplifies the effect of visual tokens on text generation, especially when the relative distance is longer between visual and text tokens. The proposed attention mechanism significantly reduces the chance of producing irrelevant text related to the video content.
Furthermore, we present a sequential visual projector that projects the current video frame into tokens of language space with the assistance of the previous frame. This approach not only captures the temporal relationship within the video, but also allows less visual tokens to encompass the entire video.
Our approach significantly outperforms various previous methods~(\emph{e.g.,} Video-ChatGPT, MovieChat) on four challenging open-ended video question answering benchmarks. 
We reach an accuracy of 60.7 on the zero-shot NExT-QA and 60.5 on the zero-shot MSRVTT-QA, setting a new state-of-the-art performance. This project is available at https://jinxxian.github.io/Vista-LLaMA.

\end{abstract}    
\section{Introduction}
\label{sec:intro}

\begin{figure}[t]
  \centering
    \includegraphics[width=1.0\linewidth]{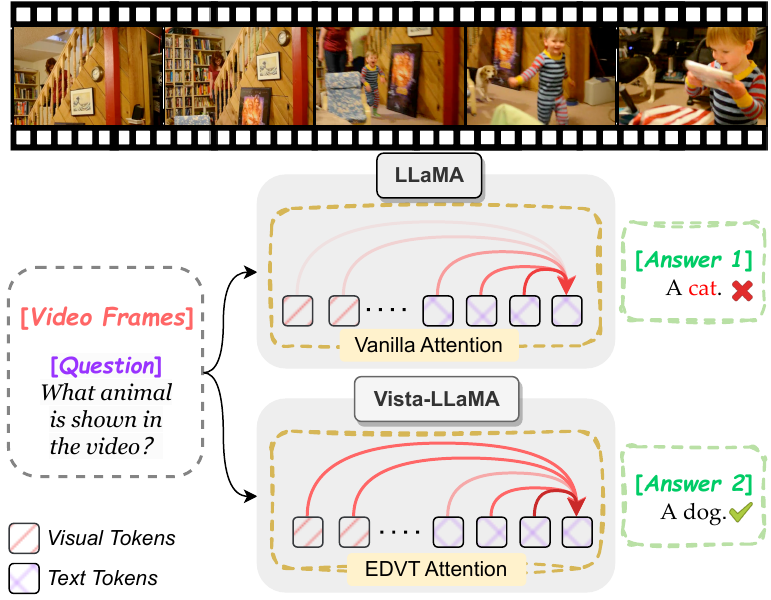}
    \caption{\textbf{Video language processing with LLaMA~\cite{Touvron2023LLaMAOA} and our \model.} The vanilla LLaMA treats visual tokens (\protect\includegraphics[scale=0.5,valign=c]{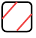}) the same as other language tokens (\protect\includegraphics[scale=0.5,valign=c]{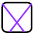}), weakening the impact for tokens in long distance. Our model retains the same mechanism for language tokens and strengthens the impact of the visual tokens. The intensity of the impact of each token is conveyed through the depth of the line color (\protect\includegraphics[scale=0.5,valign=c]{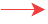}). Our model provides the accurate response for the presented scenario.}
\vspace{-12pt}
\label{fig:intro}  
\end{figure}

The surge in multi-modal vision-and-language models~\cite{alayrac2022flamingo, Li2023BLIP2BL, Liu2023VisualIT}, capable of comprehending both visual (\emph{e.g.}, image/video) and language data, can be attributed to the recent achievements of large language models (LLMs) such as GPT~\cite{floridi2020gpt}, GLM~\cite{zeng2022glm}, and LLaMA~\cite{zhang2023llama}.
Video-language models (video-LMs) pose greater challenges in scaling due to increased computational and annotation costs compared to image-language models (image-LMs). Recent research has therefore focused on effectively training video-LMs by utilizing pre-trained image-LMs~\cite{song2023moviechat, zhang2023llama}.

Video-LMs benefit from this warm-start approach to enhance visual representation learning by projecting video frames into language space and treating videos as several prompt language tokens~\cite{yang2022zero, maaz2023video}. However, this diminishes the visual impact of videos on text generation and lacks explicit temporal modeling in videos.
The generated text is often not related to the video content, as depicted in~\cref{fig:intro}, a phenomenon known as hallucination in language processing ~\cite{dhuliawala2023chain}. The distance between the generated text token and the visual tokens in large language models may be a contributing factor, especially when the visual tokens are distant from the generated text.
Additionally, handling longer videos poses a challenge because of the constraints on context length in large language models, where numerous visual tokens take up a substantial portion of the context.

We present \model, an innovative video language framework to tackle above issues. The primary concept is to preserve equal distance between all visual tokens and any language tokens, while also retaining the relative distance between any two language tokens, as depicted in \cref{fig:intro}.
\!The rotary position embedding is only applied to language tokens to capture relative distance when calculating similarity between language tokens. When computing similarity between visual and text tokens, the rotary position embedding is removed to reduce the impact of relative distance. 
With the equal distance to visual tokens (EDVT) attention, the impact of visual cues on text production is amplified without compromising the text production capability.
The experiments also show that our design produces more precise text description for input videos and the phenomenon of visual hallucinations occurring in language models has been greatly reduced.

To further improve the temporal modeling of video, we introduce a sequential visual projector.
Instead of mapping each frame of the video independently into fixed-length visual tokens, which ignores the temporal relationship between frames, we generate visual tokens for each frame using the previous projected visual tokens.
It not only incorporates the temporal relationship between frames into the language model to improve video comprehension without any extra parameters, but also enables the language model to encode longer videos with fewer visual tokens by sampling fewer projected frames.

We demonstrate the effectiveness of \model~ on four challenging video question answering~(QA) benchmarks,
where our method outperforms several previous works, and achieves the state-of-the-art in zero-shot NExT-QA~\cite{Xiao2021NExTQANP}, and MSRVTT~\cite{Xu2017VideoQA}.
We also show that our attention design and the temporal modeling mechanism improves capacity of large language model on video-question answering by a large margin. \!Comprehensive experiments are conducted to demonstrate the effectiveness of our designs.
We summarize our contributions as follows:

\begin{itemize}[leftmargin=*]
\item We introduce a novel video-language model, dubbed as \model, to enhance the video understanding and facilitate temporal modeling within the language model.
\item A novel multi-modal attention is proposed to enable reliable video text generation by maintaining equal distance to visual tokens. Additionally, temporal modeling is facilitated by employing a sequential visual projector.
\item Our method exhibits superior empirical performance, establishing new benchmarks for zero-shot open-ended video question answering tasks.
\item A detailed analysis further explains the design choices inherent in our proposed framework, contributing significantly to a better comprehension of multi-modal dynamics in large language models.
\end{itemize}
\section{Related Work}
\label{sec:related}

\subsection{Large Language Models}
Large language models (LLMs)~\cite{Scao2022BLOOMA1,Zhu2023MiniGPT4EV, OpenAI2023GPT4TR,brown2020language,Tsimpoukelli2021MultimodalFL} have demonstrated exceptional proficiency in understanding language and reasoning, resulting in the generation of high-quality natural language text in diverse domains. 
LLMs have already initiated a technological revolution and have found extensive application in various domains. Additionally, a series of open source large models, including LLaMA~\cite{Touvron2023LLaMAOA}, and OPT~\cite{Zhang2022OPTOP}, have significantly contributed to technological advancements and made remarkable contributions to the NLP community. 
Leveraging the foundation established by these impressive LLMs, researchers have further expanded their capabilities and developed exceptional models for diverse NLP tasks (\emph{e.g.} Vicuna~\cite{chiang2023vicuna}).
Our work is also built upon these remarkable LLMs, and we equip the language models with a novel attention mechanism and a sequential visual projector that enhances their abilities to comprehend visual content in videos.

\subsection{Multi-modal Large Language Models}
Researchers have been diligently exploring the application of LLMs for processing multi-modal problems~\cite{li2023videochat, Gao2022MISTM, ma2023temporal, lu2024zero}. 
The existing approaches can be classified into two primary categories. 
In the first category, LLMs are treated as controllers while the multi-modal models as tools~\cite{gupta2023visual}. 
When presented with the specific task, the LLMs first interact with user instructions and decide which tools to employ. Subsequently, it generates comprehensive responses by amalgamating the outcomes derived from these readily available multi-modal models. 
These approaches, such as Visual ChatGPT~\cite{Wu2023VisualCT}, HuggingGPT~\cite{shen2023hugginggpt}, and DoraemonGPT~\cite{yang2024doraemongpt}, have shown impressive results on various multi-modal tasks without training models.
For the second category, large-scale multi-modal models are trained on the multi-modal data.
The principle behind this line of researches is to align other modal pre-trained models with textual LLMs. Flamingo~\cite{alayrac2022flamingo} incorporates a perceiver resampler and a gated cross-attention layer to connect a frozen image encoder with the LLM. BLIP2~\cite{Li2023BLIP2BL} introduces a Q-Former to map each image into fixed-length tokens in the language embedding space via the learned queries. LLaVA~\cite{Liu2023VisualIT}, mPLUG-owl~\cite{Ye2023mPLUGOwlME}, and MiniGPT4~\cite{Zhu2023MiniGPT4EV} develop the image-LLMs utilizing image-instruction training pairs. Video Chat~\cite{li2023videochat} extends image encoders to enable large models to comprehend visual content in videos. Video-ChatGPT~\cite{maaz2023video} is trained on video instructional data to give appropriate answers for multi-modal inputs. 
In this work, we improve the understanding of videos in a video-language model by incorporating temporal modeling and a novel multi-modal attention mechanism.

\begin{figure*}[t]
  \centering
    \includegraphics[width=1.0\linewidth]{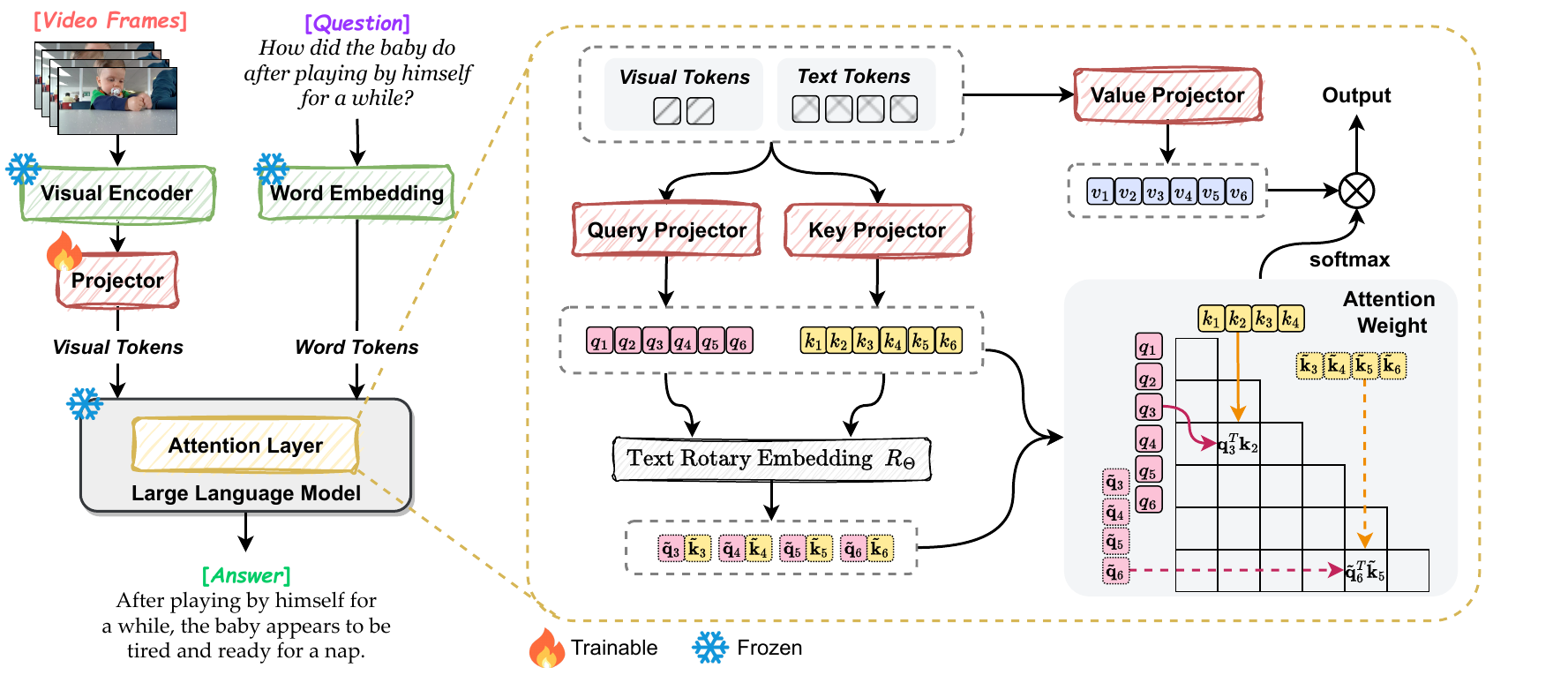}
    \caption{\textbf{The framework of \model}. The visual encoder and large language model are both frozen (\protect\includegraphics[scale=0.4,valign=c]{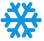}) during training, while the projector is trainable (\protect\includegraphics[scale=0.4,valign=c]{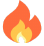}) to map video into the language's space. The attention operation in each layer is present on the right part. Only the text tokens are applied with rotary position embedding to include relative distance information. The attention weights between visual and language tokens are calculated without the rotary position embedding. The casual mask is applied to the bottom-right attention weights. }
    \label{fig:framework}  
\end{figure*}

\subsection{Video Question Answering}
Video Question Answering (VideoQA) is a task that involves answering language questions that are based on a given video~\cite{Xu2017VideoQA}. 
It requires the ability to understand and reason across different semantic levels, which in turn demands a capacity for multi-modal understanding. 
Previous VideoQA benchmarks primarily concentrated on short videos, asking questions according to the specified visual facts (\emph{e.g.}, location and objects)~\cite{Xu2016MSRVTTAL, ma2022weakly}.
Recently, several new benchmarks have been proposed to focus on resolving temporal and causal questions in longer video clips~\cite{Tapaswi2015MovieQAUS}. 
NExT-QA~\cite{Xiao2021NExTQANP} is an example of such a benchmark,  aiming to uncover the causalities or intentions of particular events, and infer subsequent actions within the video. 
In this work, we investigate the video-language model on zero-shot open-ended VideoQA where no training question-answer pairs are provided in the training stage.

\section{Method}
\label{sec:method}
\subsection{Overview}

\model~ comprises three fundamental components: a visual encoder, a visual projector, and a pre-trained large language model. \cref{fig:framework} shows an overview of \model. The components’ design and implementation details are provided below:

\begin{itemize}[leftmargin=*]
\setlength{\itemsep}{0pt}
\setlength{\parsep}{-2pt}
\setlength{\parskip}{-0pt}
\setlength{\leftmargin}{-10pt}
    \item \textbf{\textit{Visual Encoder}}. We utilize pre-trained EVA-CLIP-g~\cite{sun2023eva} and ViT-L~\cite{radford2021learning} as visual encoder. The last layer of ViT encoder is removed because it specializes in aggregating the features of the first token for contrastive learning.

    \item \textbf{\textit{Visual Projector}}. The visual projector maps the output of the visual encoder into tokens that occupy the same space as the text features from word embedding. Various visual projectors are considered in \cref{sub:svp}. The visual projector is to be trained in our work. 

    \item \textbf{\textit{Pre-trained Large Language Model}}. In this study, we utilize LLaVa~\cite{Liu2023VisualIT}, which is fine-tuned on Vicuna-7B~\cite{chiang2023vicuna} using instructional image-text pairs, for processing videos and texts. The causal mask is employed in all attention processes, incorporating the attention interplay between visual and text tokens. The pre-trained language model is frozen in the present study, and a new attention mechanism is developed to maintain a consistent distance to visual tokens for all textual tokens.
 \end{itemize}

In our framework, the video is first encoded into visual tokens using the frozen visual encoder and the trainable visual projector. These visual tokens are then combined with text tokens, which are projected with word embedding from the prompt and question sentences. The combined visual and text tokens are then input into the language model to generate answers.

\subsection{Equal Distance to Visual Tokens}
\noindent\textbf{Preliminary.}
For the concatenate visual-text input, three linear projection layers are applied in every attention layer to produce the query $\mathbf{Q}$, key $\mathbf{K}$, and value $\mathbf{V}$. Let $\mathbf{q}_{j}$ be $j^{th}$ query in $\mathbf{Q}$. 
The conventional attention mechanism updates the input by initially determining the similarity between the query and key, then applying the attention weights to the value state. Specifically, for the $j^{th}$ position, the update procedure can be articulated as follows:

\begin{equation}
\scriptsize
    \text{Attention}(\mathbf{Q}, \mathbf{K}, \mathbf{V})_{j} = \frac{\sum_{i=1}^{j}sim(\mathbf{q}_{j}, \mathbf{k}_{i})\mathbf{v}_{i}}{\sum_{i=1}^{j}sim(\mathbf{q}_{j}, \mathbf{k}_{i})},
\end{equation}
where $sim(\mathbf{q}_{j}, \mathbf{k}_{i}) = exp(\mathbf{q}_{j}^{T}\mathbf{k}_{i}/\sqrt{d})$. Here, a causal mask is utilized so that the $j^{th}$ query can only attend to the key positioned less than $j$.

\begin{algorithm}[t]
\caption{Pseudocode of EDVT attention in a PyTorch-like style.}
\label{alg:code}
\algcomment{\fontsize{7.2pt}{0em}\selectfont \texttt{qkv\_proj}: linear projection layer;  \texttt{bmm}: batch matrix multiplication; \texttt{rope}: apply rotary position embedding.
}
\definecolor{codeblue}{rgb}{0.25,0.5,0.5}
\lstset{
  backgroundcolor=\color{white},
  basicstyle=\fontsize{7.2pt}{7.2pt}\ttfamily\selectfont,
  columns=fullflexible,
  breaklines=true,
  captionpos=b,
  commentstyle=\fontsize{7.2pt}{7.2pt}\color{codeblue},
  keywordstyle=\fontsize{7.2pt}{7.2pt},
}
\begin{lstlisting}[language=python]
# x: hidden input in each attention laye
# v_mask: indcitate which input are from visual tokens 

def edvt_attention_layer(x, v_mask):
    # query, answer, value projection
    q, k, v = qkv_proj(x)   

    # apply RoPE for query and key inputs
    r_q, r_k = rope(q, k) 

    # attention weights without RoPE
    attention = bmm(q.T, k)
    
    # attention weights with RoPE
    r_attention = bmm(r_q, r_k)

    # Merge attention weights based on visual token
    attention = v_mask * attention + (1 - v_mask) * r_attention
    attention = Softmax(attention, dim=-1)

    # Update representation based on attention weights
    v = bmm(attention, v)
    out = linear_proj(v)
    return out
    
\end{lstlisting}
\end{algorithm}

\noindent\textbf{EDVT-Attention.}
The vanilla attention model lacks positional awareness, with no encoded relative distance for natural language processing. In contrast, Rotary Positional Embeddings (RoPE)~\cite{su2021roformer} encodes the position data of tokens using a rotation matrix, which inherently includes an explicit relative position dependency. Within each attention layer, RoPE is implemented across all projected query and key inputs in order to compute the attention weights via leveraging relative distance between tokens. 
The query situated at the $j^{th}$ position incorporates rotary position embedding through $\tilde{\mathbf{q}}_{j}\!=\! \mathbf{R}_{j}\mathbf{q}_{j}$, wherein $\mathbf{R}_{j}\! \in\! \mathbb{R}^{d\times d}$ represents the rotary matrix for the $j^{th}$ position. Consequently, the attention involving relative position embedding is expressed as follows:
\begin{equation}
\scriptsize
    \text{Attention}_{rope}(\mathbf{Q}, \mathbf{K}, \mathbf{V})_{j} = \frac{\sum_{i=1}^{j}sim(\mathbf{R}_{j}\mathbf{q}_{j}, \mathbf{R}_{i}\mathbf{k}_{i})\mathbf{v}_{i}}{\sum_{i=1}^{j}sim(\mathbf{R}_{j}\mathbf{q}_{j}, \mathbf{R}_{i}\mathbf{k}_{i})}.
\end{equation}

RoPE inherently integrates relative position data via the multiplication of rotation matrices rather than appending it to the input as a positional embedding. In natural language processing, this relative proximity between two words is vital, given that remote words should have less influence than adjacent words when generating the current word. 
However, using the same attention mechanism for visual and text tokens may result in unintentional text generation, a phenomenon often referred to as hallucination in LLMs. With multi-modal input, the generated text should depend on the visual content, disregarding the influence of relative distance.

To alleviate this issue and enhance the video understanding with LLMs, we introduce the EDVT-Attention where the equal distance to visual tokens is maintained while the relative distance between text tokens is attained. 
As shown in~\cref{fig:framework}, the rotary position embedding is only applied on text tokens. Let $\mathcal{V}$ and $\mathcal{T}$ be the set of visual and text tokens. The EDVT-Attention is formulated as:
\begin{equation}
\scriptsize
    \text{Attention}_{edvt}(\mathbf{Q}, \mathbf{K}, \mathbf{V})_{j} = \frac{\sum\limits_{k_i \in \mathcal{T}}sim(\tilde{\mathbf{q}}_{j}, \tilde{\mathbf{k}}_{i})\mathbf{v}_{i} + \sum\limits_{k_{i} \in \mathcal{V}}sim(\mathbf{q}_{j}, \mathbf{k}_{i})\mathbf{v}_{i}}{\sum\limits_{k_i \in \mathcal{T}}sim(\tilde{\mathbf{q}}_{j}, \tilde{\mathbf{k}}_{i}) + \sum\limits_{k_{i} \in \mathcal{V}}sim(\mathbf{q}_{j}, \mathbf{k}_{i})},
    \label{eq:edvt}
\end{equation}
where $\tilde{\mathbf{q}}_{j}$ and $\tilde{\mathbf{k}}_{j}$ denote the query and key applied with rotary embedding, separately. The distance between language tokens is determined using the rotary matrix, while the correlation between visual and textual inputs remains unaffected by the rotary embedding. This improves the influence of visual information on long-term text generation and decreases the incidence  of hallucinations wherein the fabricated content is absent in the videos. 

For ease comprehension of how EDVT-Attention works, \cref{alg:code} exhibits the pseudo-code in the decoder layer of LLMs. It involves the combinations of attention weights prior to and subsequent to the implementation of rotary embedding with the visual mask.

\begin{figure}[t]
  \centering
    \includegraphics[width=0.85\linewidth]{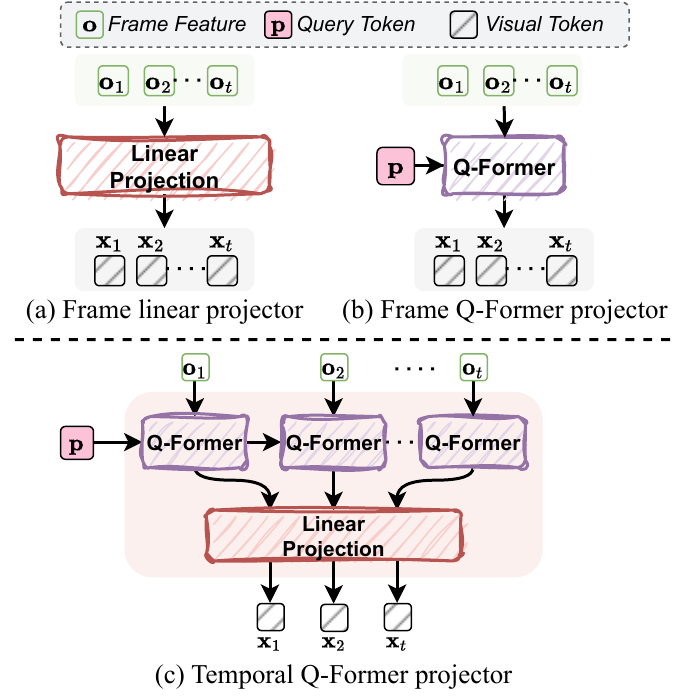}
    \caption{\textbf{Comparison of three visual projectors.} (a): Each frame feature is projected into the visual tokens independently with the linear projection. (b): Q-Former uses shared learnable query tokens to separately map each frame into fixed-length tokens. (c): The sequential Q-Former with linear projection layer to enable temporal modeling.}
    \label{fig:seq_p}  
\end{figure}

\subsection{Sequential Visual Projector}
\label{sub:svp}

The visual projector aims to map video features into the language embedding space, allowing for the fusion and processing of visual and textual inputs by the substantial language model.
As shown in~\cref{fig:seq_p}, earlier visual projectors either employ the linear layer or the query transformer (Q-Former)~\cite{Li2023BLIP2BL} to directly convert frame features into language tokens. However, the lack of temporal relationship in these methods impedes thorough video understanding in LLMs.
We introduce the sequential visual projector in~\cref{fig:seq_p} to encode the temporal context into the visual tokens. 

Let $\mathbf{o} \in \mathbb{R}^{t\times l \times d}$ be the extracted video feature of length $t$ where $l$ denotes the length of visual tokens. 
We use a Q-Former $f_{Q}$ to map each frame into the fixed length of $k$ representations $\mathbf{x}_{i} = f_{Q}(\mathbf{o}_{i}, \mathbf{p}) \in \mathbb{R}^{k \times d}$ where $\mathbf{p}\in \mathbb{R}^{k \times d}$ is the learnable query embedding.
In the prior approach, all projected frame features are merely combined with word tokens to serve as mixed input for LLMs. For encoding temporal context, we utilize the previously projected frame feature as the query to attend to the current frame feature. The current projected frame feature is then updated accordingly:
\begin{equation}
    \mathbf{x}_{t} = f_{Q}(\mathbf{o}_{i}, \mathbf{x}_{t-1}),
\end{equation}
where $\mathbf{x}_{t-1} \in \mathbb{R}^{k\times d}$ represents the previous projected frame feature. This allows the visual tokens to encode the temporal relationship, as the current frame's visual token is generated using the previous feature. 

Moreover, we can represent the entire video with fewer tokens by merely sampling a small number of visual tokens. This method tackles the challenge of encoding lengthy videos by employing sequential encoding. By integrating prior visual context into subsequent visual tokens, we can accomplish an adequate representation of the entire video through sparse sampling of projected frame features. Besides, this technique permits larger language models to manage much longer videos without length limitation.

\subsection{Implementation Details}

We fine-tune the model based on LLaVA~\cite{Liu2023VisualIT} and use 100$K$ video instruction pairs provided by~\cite{maaz2023video}.
We add Q-Former initialized from BLIP2~\cite{Li2023BLIP2BL} to project frame features into fixed-length tokens. We test our model with both ViT-L-14~\cite{radford2021learning} and EVA-CLIP-g~\cite{sun2023eva} visual encoders in experiments. 
We only update the visual projector, which contains the Q-Former and linear projection layer to project the video features to the LLMs' input space. The visual backbone and the language model are frozen during training. The \model~ is fine-tuned for 3 epochs on 8 A100 80GB GPUs with a learning rate of 2$e^{-5}$ and an overall batch size of 32. We run all the inference experiments with FP16 to save memory and faster testing.

\section{Experiments}
\label{sec:exp}

\subsection{Experimental Setup}

\noindent\textbf{Datasets.}~Our method is evaluated on four datasets:\!\!

\begin{itemize}[leftmargin=*]
\setlength{\itemsep}{0pt}
\setlength{\parsep}{-2pt}
\setlength{\parskip}{-0pt}
\setlength{\leftmargin}{-10pt}
    \item \textbf{NExT-QA}~\cite{Xiao2021NExTQANP} is designed to advance video understanding from descriing to explaining the temporal actions. It comprises 5,440 videos and approximately 52$K$ manually annotated QA pairs, which are categorized into \textit{temporal} (Tem.), \textit{causal} (Cau.), and \textit{descriptive} (Des.) questions. 

    \item \textbf{MSVD-QA}~\cite{Xu2017VideoQA} is a dataset built upon Microsoft Research Video Description Corpus~\cite{Chen2011CollectingHP}, commonly used in video caption tasks. The MSVD-QA dataset comprises a total of 1,970 video clips with 50,505 QA pairs.

    \item \textbf{MSRVTT-QA}~\cite{Xu2017VideoQA} is based on MSR-VTT dataset~\cite{Xu2016MSRVTTAL}, which includes 10$K$\! videos and 243$K$\! QA pairs with larger and has more complex scenes. 

    \item \textbf{ActiviytNet-QA}~\cite{Yu2019ActivityNetQAAD} is a fully annotated and large-scale videoQA dataset. It contains 58$K$ QA pairs derived from 5,800 complex web videos derived from the popular ActivityNet dataset~\cite{Heilbron2015ActivityNetAL}.

\end{itemize}

\noindent\textbf{Evaluation.}
We adopt two metrics to evaluate the performance of video-language models.

\begin{itemize}[leftmargin=*]
\setlength{\itemsep}{0pt}
\setlength{\parsep}{-2pt}
\setlength{\parskip}{-0pt}
\setlength{\leftmargin}{-10pt}

\item \textbf{Open-Ended Zero-Shot Question-Answer Evaluation.}  
 As video-language models generate responses of varying lengths to open-ended questions, it is challenge to evaluate models with traditional word matching strategy. We employ LLM-Assisted evaluation, in line with~\cite{maaz2023video}, for fair comparison. Given the question, correct answer, and predicted answer, GPT3.5-turbo-0613 is used to return \textit{True} or \textit{False} judgement and relative score ($0$ to $5$). 

\item \textbf{Video-Based Text Generation Benchmarking.}
We evaluate the text generation performance following~\cite{maaz2023video} from five aspects: \textit{Correctness of Information}, \textit{Consistency}, \textit{Detail Orientation}, \textit{Contextual Understanding},  and \textit{Temporal Understanding}.
The test set for this evaluation is based on the ActivityNet-200~\cite{caba2015activitynet}, featuring videos with rich, dense descriptive captions and associated question-answer pairs from human annotations.
The evaluation pipeline is also built with the GPT-3.5 model and relative score ($0$ to $5$) is generated. 

\end{itemize}

\begin{table*}[t]
\centering
\small
\setlength{\tabcolsep}{10pt}
\renewcommand\arraystretch{1.1}
\resizebox{0.95\linewidth}{!}{
\begin{tabular}{l|| c c| c c| c c| c c}
\hline \thickhline
\rowcolor{mygray}
\textbf{Method}  & \multicolumn{2}{c|}{\textbf{NExT-QA~\cite{Xiao2021NExTQANP}}} & \multicolumn{2}{c|}{\textbf{MSVD-QA~\cite{Xu2017VideoQA}}} & \multicolumn{2}{c|}{\textbf{MSRVTT-QA~\cite{Xu2017VideoQA}}} & \multicolumn{2}{c}{\textbf{ActivityNet-QA~\cite{Yu2019ActivityNetQAAD}}} \\
\cline{2-9}
\rowcolor{mygray}
  &  \textbf{Accuracy} & \textbf{Score} &  \textbf{Accuracy} & \textbf{Score} & \textbf{Accuracy} & \textbf{Score} & \textbf{Accuracy} & \textbf{Score}\\
\hline
\hline
FrozenBiLM~\cite{yang2022zero}    &- &- & 32.2 & - & 16.8 & - & 24.7 & - \\
Video Chat~\cite{li2023videochat}  & \underline{56.2} & \underline{3.2}  & 56.3 & 2.8 & 45.0 & 2.5 & 26.5 & 2.2 \\
LLaMA Adapter~\cite{zhang2023llama}  & - & - & 54.9 & 3.1 & 43.8 & 2.7 & 34.2 &\underline{2.7} \\
Video LLaMA~\cite{zhang2023video} & - & - & 51.6 & 2.5 & 29.6 & 1.8 & 12.4 & 1.1 \\
MovieChat~\cite{song2023moviechat}  & 49.9 & 2.7 & 61.0 & 2.9 &\underline{49.7} &\underline{2.8} &\textbf{51.5} & 3.1\\ 
Video-ChatGPT~\cite{maaz2023video}  & 54.6 &\underline{3.2} & \underline{64.9} & \underline{3.3} & 49.3 & \underline{2.8}  &35.2 & \underline{2.7} \\ 
\midrule
\textbf{\model}~(\textbf{Ours})  &\textbf{60.7} &\textbf{3.4} &\textbf{65.3} &\textbf{3.6}  &\textbf{60.5} &\textbf{3.3} &\underline{48.3}  &\textbf{3.3} \\   
\hline \thickhline

\end{tabular}}
\caption{\textbf{Comparison with SoTA methods} on zero-shot VideoQA. See \S\ref{sec:sota} for more details.}
\label{tab:comp_sota}
\end{table*}
\subsection{Comparison to State-of-the-Arts}
\label{sec:sota}
For the zero-shot open-ended video question answering tasks, we compare our model with FrozenBiLM~\cite{yang2022zero}, Video Chat~\cite{li2023videochat}, LLaMA Adapter~\cite{zhang2023llama}, VideoLLaMA~\cite{zhang2023video}, Video-ChatGPT~\cite{maaz2023video}, and MovieChat~\cite{song2023moviechat}  in \cref{tab:comp_sota}. 
FrozenBiLM adapts frozen the bidirectional language model, showing promising results in zero-shot VideoQA settings. Other compared models are built on recent large auto-regressive language models.
Despite pre-existing models have produced substantial results, \model~ consistently outperforms them, achieving state-of-the-art (SoTA) performance across three datasets: NExT-QA~\cite{Xiao2021NExTQANP}, MSVD-QA~\cite{Xu2017VideoQA}, and MSVTT-QA~\cite{Xu2017VideoQA}. Our method obtains the highest results, with 60.5\% accuracy on MSRVTT, elevating the performance of the second-best model by nearly 10\%. Additionally, our method attains 60.7\% accuracy on NExT-QA, markedly superior to Video-ChatGPT~\cite{maaz2023video}. These results demonstrate \model's capability to comprehend video content and produce precise answers.

\begin{table}[t]
\centering
\small
\setlength{\tabcolsep}{8pt}
\renewcommand\arraystretch{1.1}
\resizebox{0.48\textwidth}{!}{
\begin{tabular}{l||c|c|c|c|c}
\hline \thickhline

\rowcolor{mygray} 
\textbf{Method} &\textbf{Cr.} &\textbf{Cs.}  &\textbf{De.} &\textbf{Ct.} &\textbf{Te.} \\
\hline
\hline

Video Chat~\cite{li2023videochat} &2.23 &2.24 &2.50 &2.53 &1.94 \\
LLaMA Adapter~\cite{zhang2023llama} &2.03 &2.15 &2.32 &2.30 &\underline{1.98} \\
Video LLaMA~\cite{zhang2023video} &1.96 &1.79 &2.18 &2.16 &1.82 \\
Video-ChatGPT~\cite{maaz2023video} &\underline{2.40} &\textbf{2.37} &\underline{2.52} &\underline{2.62} &\underline{1.98} \\
\midrule
\textbf{\model~(Ours)} &\textbf{2.44} &\underline{2.31} &\textbf{2.64} &\textbf{3.18} &\textbf{2.26} \\

\hline \thickhline

\end{tabular}}
\caption{\textbf{Quantitative results} on video-based text generation with different video-language methods (\S\ref{sec:sota}). For clarity, five scores are reported (``Cr.'': \textit{Correctness of Information}, ``Cs.'': \textit{Consistency}, ``De.'': \textit{Detail Orientation}, ``Ct'': \textit{Contextual Understanding}, ``Te.'': \textit{Temporal Understanding}).}
\label{tab:comp_gen}
\end{table}

We present the results of the evaluation of video-based text generation in \cref{tab:comp_gen}. 
The results reveal its competent performance across all aspects when compared with the recent video-language models, Video Chat\cite{li2023videochat}, VideoLLaMA~\cite{zhang2023video}, and Video-ChatGPT~\cite{maaz2023video}. 
Despite being trained with identical datasets, our model outperforms Video-ChatGPT in four aspects. Our mothod offers a more comprehensive interpretation, and its responses are more in tune with the overarching context of the video content than comparable approaches. By utilizing the EDVT-Attention and sequential temporal modeling techniques, our model demonstrates a strong ability to generate text that is contextually appropriate, detailed, and includes precise timing for video inputs.

\subsection{Abalation Study}
\label{sec:ablation}

\begin{figure}
    \centering
    \includegraphics[width=0.8\linewidth]{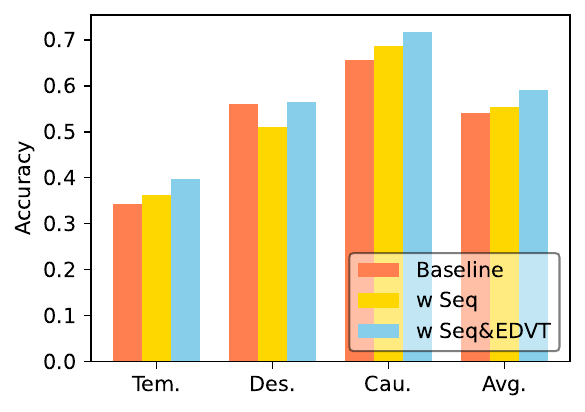}
    \caption{\textbf{Comparison of different design choices} on NExT-QA~\cite{Xiao2021NExTQANP}(\S\ref{sec:ablation}). For clarity, accuracy of base model and two variants are given (``Baseline'': the frozen LLaVA~\cite{Liu2023VisualIT} with trainable Q-Former~\cite{Li2023BLIP2BL}, ``\textit{w Seq}'': base model with sequential visual projector, ``\textit{w Seq\&EDVT}'': base model with both sequential visual projector and EDVT-Attention).}
    \label{fig:comp_seq_edvt}
\end{figure}

\noindent\textbf{Effect of Design Choices.}~We investigate the effect of our two designs, including the equal-distance to visual-tokens (EDVT) attention and the sequential visual projector.
As shown in \cref{fig:comp_seq_edvt}, our model is tested on NExT-QA~\cite{Xiao2021NExTQANP}, which comprises three different types of questions. 
The baseline model employs LLaVA~\cite{Liu2023VisualIT} as the language model, and Q-Former in BLIP-2~\cite{Li2023BLIP2BL} as the visual projector. We first incorporate temporal modeling into the baseline, utilizing the projected tokens from the previous frame as the query tokens to generate visual tokens for the current frame, a setup denoted as ``\textit{w Seq}" in \cref{fig:comp_seq_edvt}. 
Further enhancing this version with EDVT-Attention, we introduce ``\textit{w Seq\&EDVT}", ultimately forming our final \model. The results affirm the efficacy of our design across all question types.

\noindent\textbf{EDVT-Attention.} We here validate the effectiveness of our core EDVT-Attention. Table~\ref{tab:comp_edvt} reports the comparison results of different models in combination with our EDVT-Attention on NExT-QA~\cite{Xiao2021NExTQANP}. 
Initially, we integrate the EDVT-Attention into Video-ChatGPT~\cite{maaz2023video}, where only the attention component is substituted with our EDVT-Attention during training.
Further deployment of the EDVT-Attention into Video-ChatGPT~\cite{maaz2023video} yields significant performance gains (\emph{e.g.,} 65.1\% escalates to 72.8\% for causal questions).
The accuracy is enlarged across all three setting types, indicating that our EDVT-Attention considerably enhances the multi-modal understanding in large language models.

\begin{figure}
    \centering

    \begin{subfigure}[b]{0.48\linewidth}
        \includegraphics[width=\textwidth]{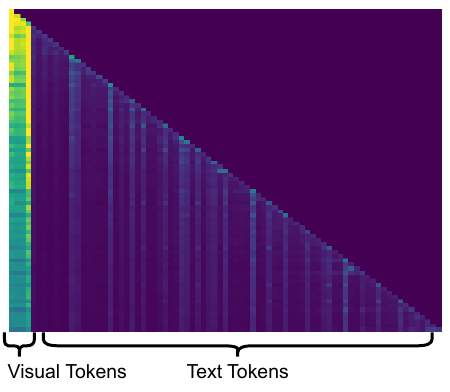}
        \caption{Vanilla attention}
        \label{fig:sub1}
    \end{subfigure}
    \begin{subfigure}[b]{0.48\linewidth}
        \includegraphics[width=\textwidth]{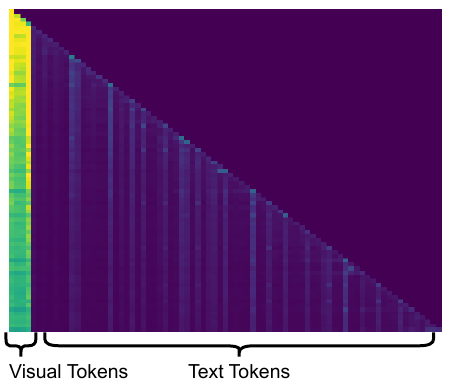}
        \caption{EDVT-Attention (Ours)}
        \label{fig:sub2}
    \end{subfigure}
    \caption{\textbf{Comparison of attention weights} for varing context lengths. Lighter colors represent higher weights. To improve clarity, we have combined visual token weights into the first four tokens. We recommend zooming in for optimal viewing.}
    \label{fig:attention}
\end{figure}

We further illustrate the attention weights of both EDVT-Attention and conventional attention in \cref{fig:attention}. For this experiment, we employ 128 visual tokens and integrate the attention weights into the first four tokens to enhance visualization. The attention weights assigned to visual tokens are markedly greater in our EDVT-Attention compared to those in conventional attention. This indicates that the impact of visual tokens on language tokens is considerably more substantial under the proposed EDVT-Attention, shedding light on why the performance notably improves with our design.

\begin{table}
\small

\resizebox{0.48\textwidth}{!}{

\setlength\tabcolsep{2pt}
\renewcommand\arraystretch{1.1}

\begin{tabular}{l||cccc}
\hline \thickhline
\rowcolor{mygray} 
\multicolumn{1}{l||}{\cellcolor{mygray}}   &\multicolumn{4}{c}{\cellcolor{mygray}\textbf{NExT-QA~\cite{Xiao2021NExTQANP}}}  \\     \cline{2-5} 
\rowcolor{mygray} 
\multicolumn{1}{l||}{\multirow{-2}{*}{\cellcolor{mygray}\textbf{Method}}} &\textbf{Tem.}
&\multicolumn{1}{c}{\cellcolor{mygray}\textbf{Cau.}} & \multicolumn{1}{c}{\cellcolor{mygray}\textbf{Des.}} & \multicolumn{1}{c}{\cellcolor{mygray}\textbf{Avg.}} \\ 
\hline
\hline

\multicolumn{1}{l||}{Video-ChatGPT~\cite{maaz2023video}}  & 37.6	& 65.1 & 54.9 & 54.6 \\
+~\textbf{\texttt{EDVT-Attention}}  & 39.5~\color{mygreen}{({+1.9})} & 72.8~\color{mygreen}{({+7.7})} & 54.8 & 59.3~\color{mygreen}{({+4.7})} \\ 
\midrule

\model~(ViT-L-14) & 34.0 & 69.1 & 42.2 & 53.6  \\
+~\textbf{\texttt{EDVT-Attention}} & 36.8~\color{mygreen}{({+2.8})} & 72.2~\color{mygreen}{({+3.1})} & 47.7~\color{mygreen}{({+5.5})} & 56.5~\color{mygreen}{({+2.9})}  \\ 
\midrule

\model~(EVA-CLIP-g)  & 34.3 & 65.8 & 55.9 & 54.1 \\
+~\textbf{\texttt{EDVT-Attention}}  & 40.7~\color{mygreen}{({+6.4})} & 72.3~\color{mygreen}{({+6.5})} & 57.0~\color{mygreen}{({+1.1})} & 59.7~\color{mygreen}{({+5.6})} \\ 
\hline \thickhline
\end{tabular}}
\caption{\textbf{Comparison of EDVT-Attention design} with diffent visual encoders on NExT-QA~\cite{Xiao2021NExTQANP} (\S\ref{sec:ablation}). For clarity, accuracy of all questions and three types of questions are reported (``Tem.'': \textit{temporal}, ``Cau.'': \textit{causal}, ``Des.'': \textit{descriptive}, ``Avg.'': average).}

\label{tab:comp_edvt}
\end{table}


\begin{table}
\small

\resizebox{0.48\textwidth}{!}{

\setlength\tabcolsep{10pt}
\renewcommand\arraystretch{1.1}

\begin{tabular}{l||cccc}
\hline \thickhline
\rowcolor{mygray} 
\multicolumn{1}{l||}{\cellcolor{mygray}}   &\multicolumn{4}{c}{\cellcolor{mygray}\textbf{NExT-QA~\cite{Xiao2021NExTQANP}}}  \\     \cline{2-5} 
\rowcolor{mygray} 
\multicolumn{1}{l||}{\multirow{-2}{*}{\cellcolor{mygray}\textbf{Visual projector}}} &\textbf{Tem.}
&\multicolumn{1}{c}{\cellcolor{mygray}\textbf{Cau.}} & \multicolumn{1}{c}{\cellcolor{mygray}\textbf{Des.}} & \multicolumn{1}{c}{\cellcolor{mygray}\textbf{Avg.}} \\ 
\hline \hline

Linear Projector  & 37.6	& 65.1 & 54.9 & 54.6 \\
Q-Former (\textit{BERT init.}) & 35.2 & 62.7 & 49.2 & 51.8 \\ 
Q-Former (\textit{BLIP-2 init.}) & 34.3 & 65.8 & 55.9 & 54.1 \\
SeqQ-Former (\textit{BLIP-2 init.}) & 36.2 & 68.5 & 51.1 & 55.4 \\
\bottomrule
\end{tabular}}
\caption{\textbf{Comparison of different visual projectors} on NExT-QA~\cite{Xiao2021NExTQANP}. The linear projector is initialized with pre-trained weights in LLaVa~\cite{Liu2023VisualIT}. \textit{BERT init.} and \textit{BLIP-2 init.} indicate that the visual projector is initialized with weights from BERT~\cite{devlin2018bert} and BLIP-2~\cite{Li2023BLIP2BL}. SeqQ-Former is the proposed sequential visual projector.  See \S\ref{sec:ablation} for more details.}
\label{tab:com_qformer}
\end{table}

\noindent\textbf{Visual Projector.}
The visual projector is trainable within the model. Therefore, we have examined the impacts of different visual projector designs, as shown in Table~\ref{tab:com_qformer}.
We evaluated three visual projector variants. The Q-Former, initialized using BLIP-2~\cite{Li2023BLIP2BL}, considerably outperforms the Q-Former that was initialized with BERT~\cite{devlin2018bert}.
This indicates that visual projectors, when pre-trained on image-text pairs, can significantly enhance video comprehension. The model's performance peaks when integrating the sequential design, referred to as ``SeqQ-Former'' in the figure, outperforming all other visual projectors in terms of accuracy.

\begin{figure}
    \centering
    \includegraphics[width=\linewidth]{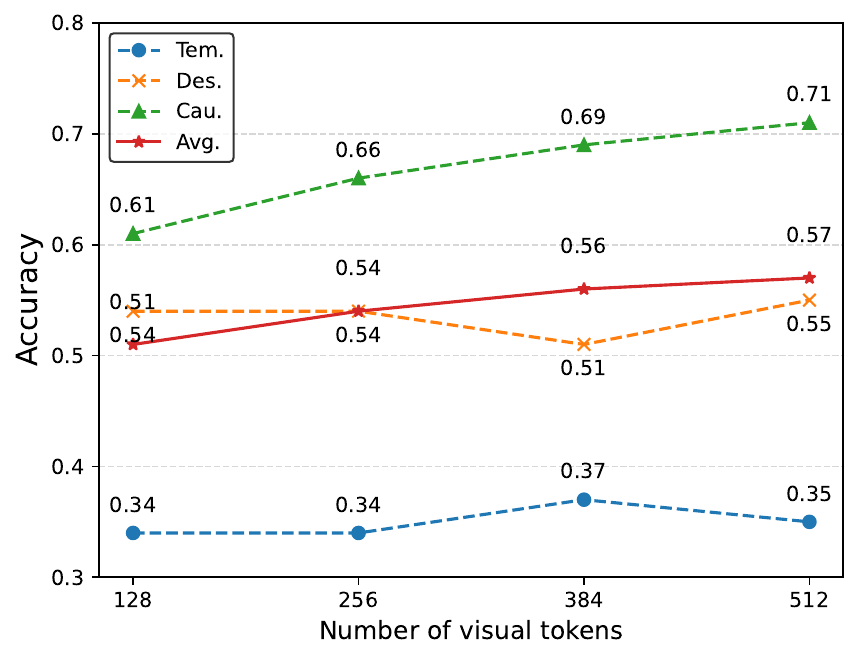}
    \caption{\textbf{The effect of training visual tokens} on NExT-QA~\cite{Xiao2021NExTQANP}. Accuracy of all questions and three types of questions, including \textit{temporal} (Tem.), \textit{descriptive} (Des.), and \textit{causal} (Cau.), are presented with different colors.}
\label{fig:frame}
\end{figure}

\begin{figure*}
    \centering
    \includegraphics[width=\linewidth]{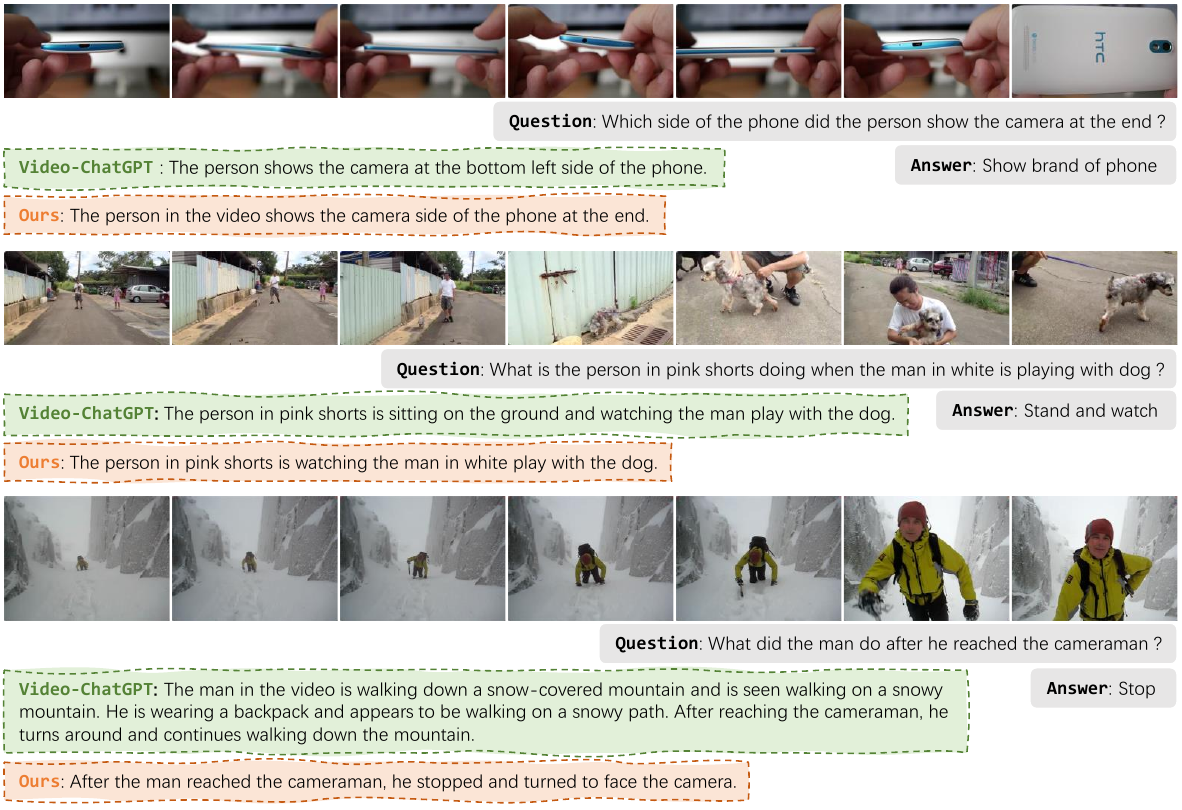}
    \caption{\textbf{Visualization results} on different video questions. The questions and annotated answers are located on the left side. The generated text from Video-ChatGPT~\cite{Xu2023BaizeAO} and our model is presented in the green and orange boxes, respectively. See \S\ref{sec:ablation} for more details.}
    \label{fig:visual}
\end{figure*}

\noindent\textbf{Number of Training Visual Tokens.}
We assess the impact of visual tokens on NExT-QA as illustrated in \cref{fig:frame}. In this study, a frame is converted into 32 visual tokens using Q-Former. Frames were sampled at various timestamps throughout the training process. The overall accuracy improves as more visual tokens are incorporated into the language model. However, for temporal queries, precision declined with an increased number of visual tokens. The accuracy of temporal reasoning questions is notably lower than that of other question types. Temporal reasoning presents more difficulties, and the language model may not excel at the temporal modeling of visual tokens. In terms of descriptive question types, the accuracy steadily increases as the model gains more visual information.

\noindent\textbf{Quantitative Results.}
We visualize the generated responses of Video-ChatGPT~\cite{maaz2023video} and \model~ for different videos in \cref{fig:visual}. 
The frames sampled at different timestamps is presented, with questions listed below the images. In the first video, the mobile phone is turned, sequentially exposing each side over time. The question asks for the side of the mobile phone at the end of the video. Video-ChatGPT gives the wrong response, as the camera is never present at the bottom left side in the video. The hallucination occurs in Video-ChatGPT, whereas our model predicts the answer correctly.
The responses in three cases demonstrate that Video-ChatGPT provides unrelated answers that do not correspond to the video content. 
In the second case, the person in pink is always standing while the man is sitting on the ground, but Video-ChatGPT incorrectly states that the person in pink is also sitting on the ground.
In the third video, the man is consistently walking up the mountain and never walking down. 
However, Video-ChatGPT falsely outputs that the man continues walking down the mountain.
In contrast, our model provides the correct and reliable responses. This demonstrates that our model significantly reduces hallucination in video understanding and delivers more accurate responses.

\section{Conclusion}
In this work, we present \model ~to improve the video understanding in large language model. A new vision-aware attention is introduced to maintain same relative distance between all visual tokens and language tokens. 
In addition, we propose a sequential visual projector to map video into the language space to enable temporal modeling. 
Experiments on several video question answering task demonstrate that our designs significant improves the current SoTA. 
Current work is built on the pre-trained image-text architecture, the power of our designs could be enlarged when applied in pre-training stage.  
Furthermore, the performance of our sequential encoder on longer videos can also be assessed.
 We would consider evaluating the model in more tasks and applied the attention mechanism to more vision language models.

\noindent\textbf{Acknowledgements.} This work was supported by the National Natural Science Foundation of China (62293554, U2336212), and China Postdoctoral Science Foundation (524000-X92302).

{
    \small
    \bibliographystyle{ieeenat_fullname}
    \bibliography{main}
}

\clearpage
\setcounter{page}{1}
\maketitlesupplementary

\section{Additional Results}
\label{sec:rationale}

\begin{figure*}
    \centering
    \includegraphics[width=\linewidth]{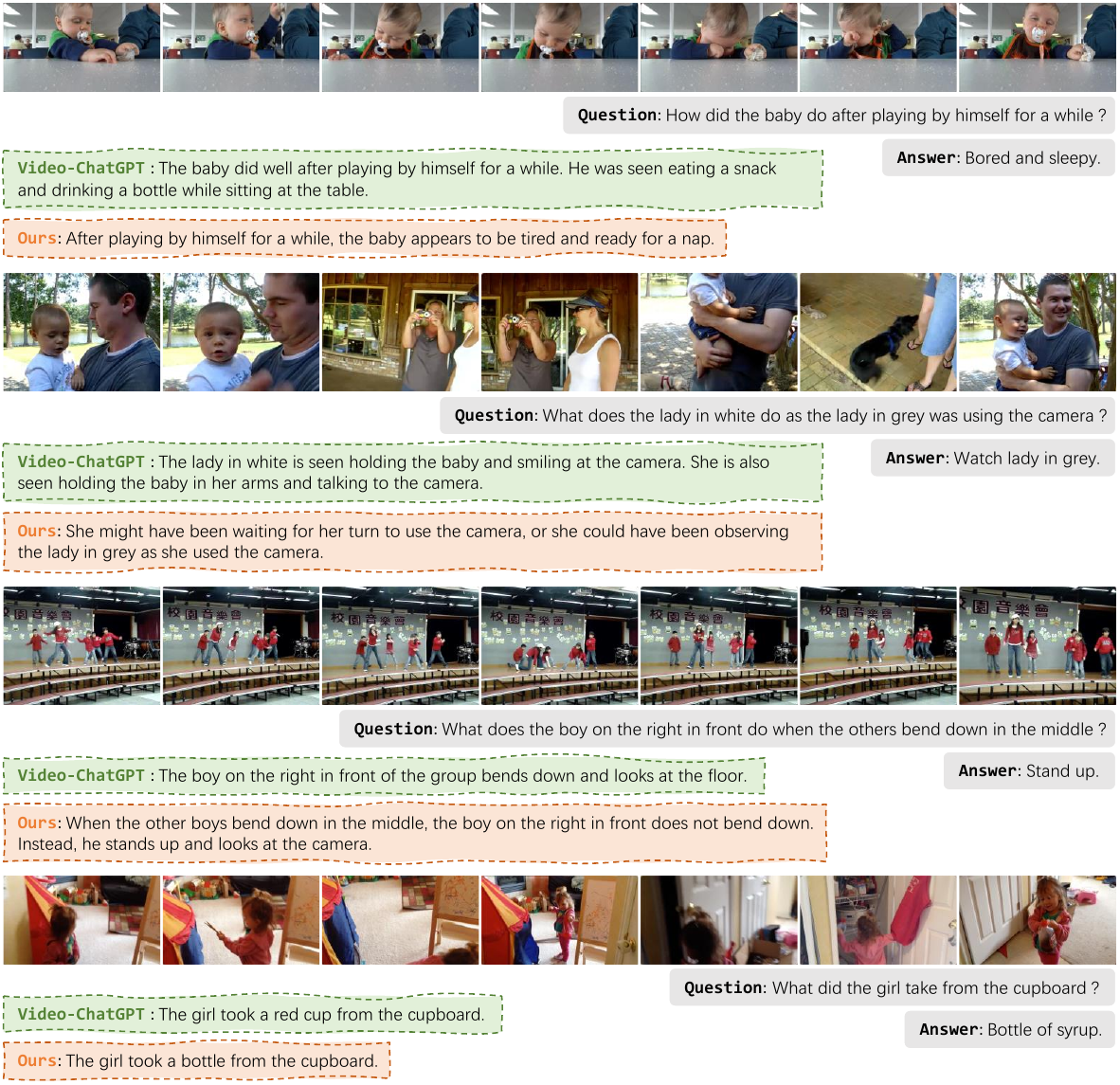}
    \caption{\textbf{Visualization results} on different video questions.}
    \label{fig:supp_visual_1}
\end{figure*}

\begin{figure*}
    \centering
    \includegraphics[width=\linewidth]{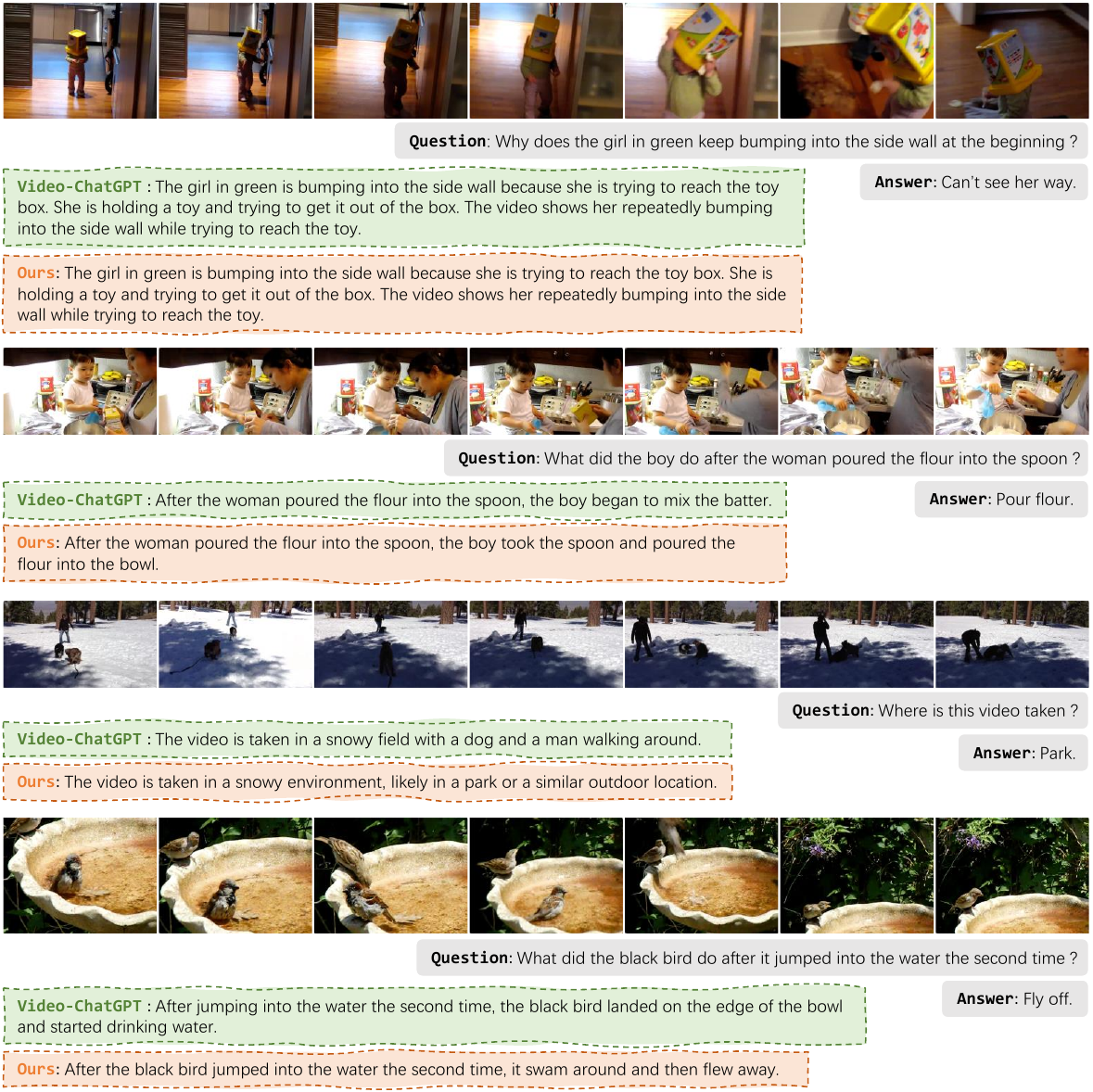}
    \caption{\textbf{Visualization results} on different video questions.}
    \label{fig:supp_visual_2}
\end{figure*}

\noindent\textbf{Visualization Results.}
We present additional visualization results for different video questions in \cref{fig:supp_visual_1,fig:supp_visual_2}. In comparison to Video-ChatGPT~\cite{maaz2023video}, our \model~provides more reasonable answers and descriptions that align better with the video content. Video-ChatGPT often responds with irrelevant information, resulting in hallucinations. For example, in the first video where the baby appears tired, Video-ChatGPT incorrectly states that the baby was eating a snack, even though there is no eating action shown in the video. We have more examples that demonstrate the improvement of our method on NExT-QA~\cite{Xiao2021NExTQANP}. We only showcase a small portion of these cases to highlight the differences. Our methods achieve much better performance and offer more reliable replies due to the proposed EDVT-Attention, which maintains an equal distance to all visual tokens.

\begin{figure*} 
    \centering

    \begin{subfigure}{0.9\textwidth}
        \includegraphics[width=\linewidth]{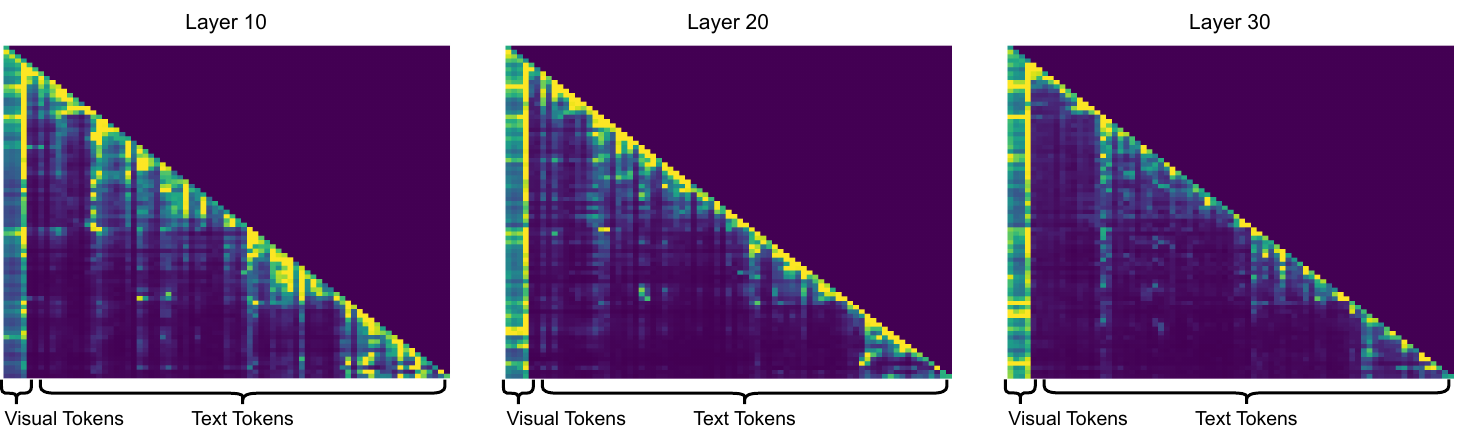}
        \caption{Vanilla attention}
        \label{fig:sub1}
    \end{subfigure}

    \medskip  

    \begin{subfigure}{0.9\textwidth}
        \includegraphics[width=\linewidth]{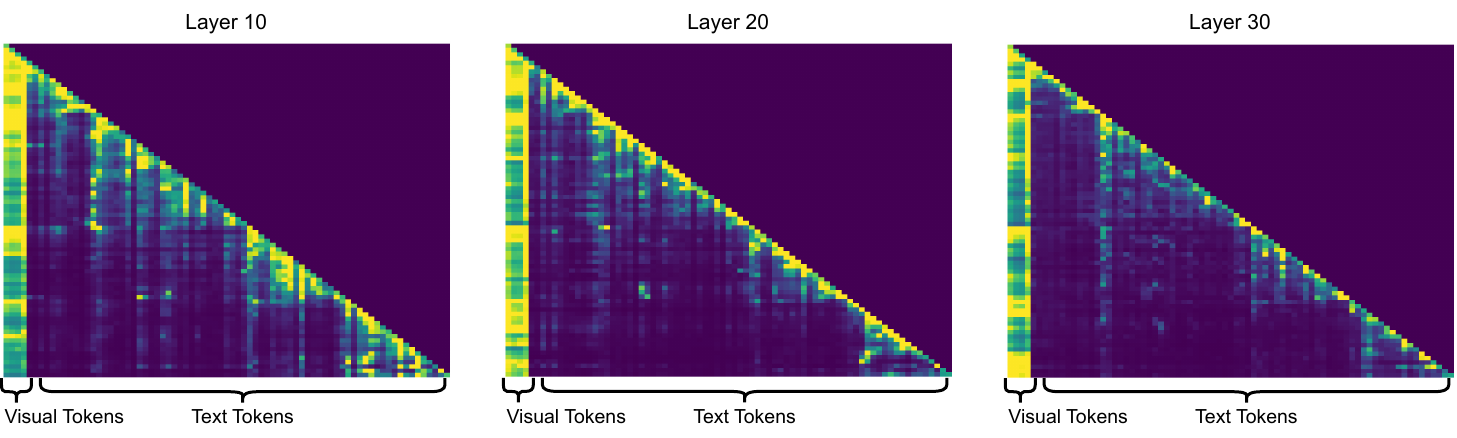}
        \caption{EDVT-Attention (Ours)}
        \label{fig:sub4}
    \end{subfigure}

\caption{\textbf{Comparison of attention weights} for varing context lengths in different layers. Lighter colors represent higher weights. To improve clarity, we have combined visual token weights into the first four tokens. We recommend zooming in for optimal viewing.}

\label{fig:supp_attention}
\end{figure*}

\noindent\textbf{Attention Weights in Different Layers.}
In \cref{fig:supp_attention}, we present the attention weights in different layers. Different from \cref{fig:attention}, here we sum instead of average the attention weights of 32 heads to present clear comparison. From the figure, we show that the attention weights between text tokens in the EDVT-Attention are larger than attention weights in Vanilla attention. It indicates that the EDVT-Attention strengthen the impact of visual tokens on generating text.

In \cref{fig:supp_attention}, we visualize the attention weights in different layers. In contrast to \cref{fig:attention}, where we averaged the attention weights of 32 heads, here we present the sum for a clearer comparison. The figure reveals that the attention weights between text tokens in the EDVT-Attention are greater than those in Vanilla attention. This suggests that the EDVT-Attention enhances the influence of visual tokens on text generation.

\begin{table}
\small

\resizebox{0.48\textwidth}{!}{

\setlength\tabcolsep{10pt}
\renewcommand\arraystretch{1.1}

\begin{tabular}{l||l||cccc}
\hline \thickhline
\rowcolor{mygray} 
\multicolumn{1}{l||}{\cellcolor{mygray}} & \multicolumn{1}{l||}{\cellcolor{mygray}}   &\multicolumn{4}{c}{\cellcolor{mygray}\textbf{NExT-QA~\cite{Xiao2021NExTQANP}}}  \\     

\cline{3-6} 
\rowcolor{mygray} 
\multicolumn{1}{l||}{\multirow{-2}{*}{\cellcolor{mygray}\textbf{Query}}} 
&\multicolumn{1}{l||}{\multirow{-2}{*}{\cellcolor{mygray}\textbf{Key}}} 
&\textbf{Tem.}
&\multicolumn{1}{c}{\cellcolor{mygray}\textbf{Cau.}} & \multicolumn{1}{c}{\cellcolor{mygray}\textbf{Des.}} & \multicolumn{1}{c}{\cellcolor{mygray}\textbf{Avg.}} \\ 
\hline \hline

RoPE  & RoPE  & 34.3 & 65.8 & 55.9 & 54.1 \\
FixVPE & FixVPE & 37.0 & 70.5 & 56.7 & 57.6 \\
RoPE & EDVT & 32.2 & 48.1 & 41.8 & 42.0 \\ 
EDVT & EDVT & 40.7 & 72.3 & 57.0 & 59.7 \\
\bottomrule
\end{tabular}}
\caption{\textbf{Comparison of positional embedding strategies} on NExT-QA~\cite{Xiao2021NExTQANP}.
We provide a list of various positional embedding strategies used for query and key vectors in the attention layer. The ``RoPE" indicates the use of rotary positional embedding for all visual and text tokens. The ``FixVPE" refers to the fixed position rotary positional embedding used for all visual tokens. Lastly, ``EDVT" indicates that the rotary positional embedding is exclusively applied to text tokens.}
\label{tab:com_posemb}
\end{table}

\noindent\textbf{Positional Embedding Study.}
We explored various strategies for positional embedding in the attention layer, focusing on the query and key vectors. According to \cref{tab:com_posemb}, the model achieves the highest accuracy when only text tokens have rotary positional embedding applied to both the query and key vectors. When only the query vectors have RoPE applied and the key vectors do not, the performance decreases significantly. This is because the relative distance is compromised when only the query has RoPE. We also attempted to use fixed positional embedding on all visual tokens. Unlike in DEVT, all visual tokens have RoPE applied with the same position index of 0. Compared to the baseline, this modification also improves performance on different question types. However, it is still inferior to our design. This demonstrates that the proposed EDVT design truly enhances video understanding in LLMs.

\section{Movie Evaluation}

\begin{table*}[t]
\centering
\small
\setlength{\tabcolsep}{10pt}
\renewcommand\arraystretch{1.1}
\resizebox{0.95\linewidth}{!}{
\begin{tabular}{l|| c c| c c| c c| c c| c c| c c}
\hline \thickhline
\rowcolor{mygray}
\textbf{Method}  & \multicolumn{2}{c|}{\textbf{Overall}} & \multicolumn{2}{c|}{\textbf{Description}} & \multicolumn{2}{c|}{\textbf{Temporality}} & \multicolumn{2}{c|}{\textbf{Spaciality}} & \multicolumn{2}{c|}{\textbf{Intention}} & \multicolumn{2}{c}{\textbf{Perception}} \\
\cline{2-13}
\rowcolor{mygray}
  & \textbf{Score} &  \textbf{Accuracy} & \textbf{Score} &  \textbf{Accuracy} & \textbf{Score} & \textbf{Accuracy} & \textbf{Score} & \textbf{Accuracy} & \textbf{Score} & \textbf{Accuracy} & \textbf{Score} & \textbf{Accuracy} \\
\hline
\hline
MovieChat & 2.11 & 20.86 & 2.41 & 23.67 & 1.97 & 16.32 & 1.98 & 16.40 & 2.41 & 30.19 & 1.97 & 21.80 \\
Video-LLAMA & 2.27 & 23.17 & 2.31 & 19.30 & 2.12 & 16.35 & 2.19 & 21.95 & 2.47 & 31.94 & 2.35 & 27.70 \\
Video-ChatGPT & 2.60 & 34.11 & 2.55 & 26.24 & 2.60 & 34.11 & 2.50 & 30.62 & 2.94 & 46.36 & 2.43 & 31.77 \\
\midrule
\textbf{\model}~(\textbf{Ours})  &\textbf{2.98} &\textbf{44.90} &\textbf{2.79} &\textbf{31.46} &\textbf{2.92} &\textbf{46.22} &\textbf{2.73} &\textbf{35.63} &\textbf{3.38} &\textbf{61.89} &\textbf{3.12} &\textbf{47.49} \\   
\hline \thickhline

\end{tabular}}
\caption{\textbf{Performance Comparison on CineClipQA} of different methods on various classifications. }
\label{tab:comp_algocmps}
\end{table*}
\begin{figure*}
    \centering
    \includegraphics[width=\linewidth]{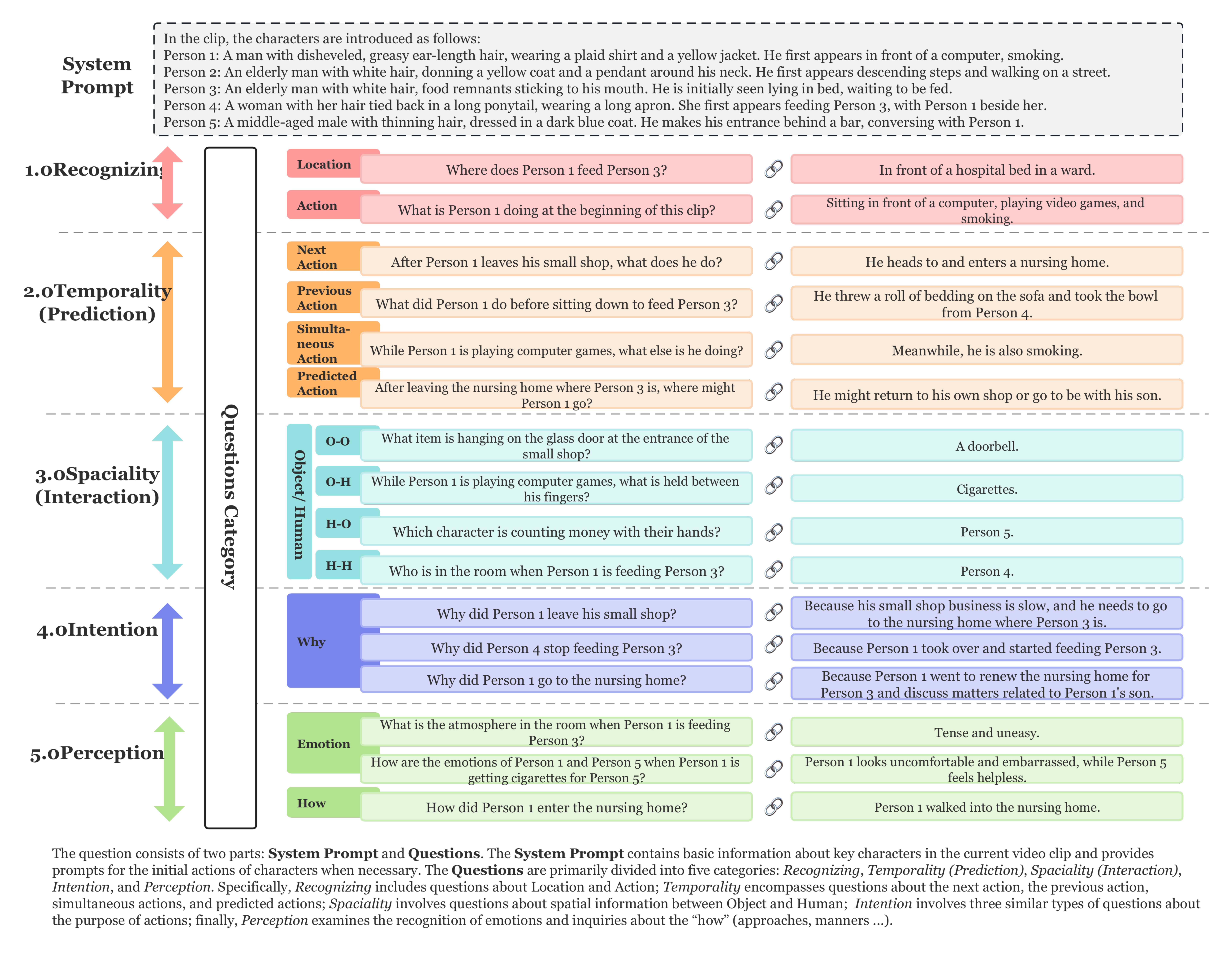}
    \caption{\textbf{CineClipQA}, a novel dataset meticulously crafted to probe the capabilities of visual language models in comprehending and interpreting plot-driven video content.}
\label{fig:ccdes}
\end{figure*}
\begin{figure*}
    \centering
    \includegraphics[width=\linewidth]{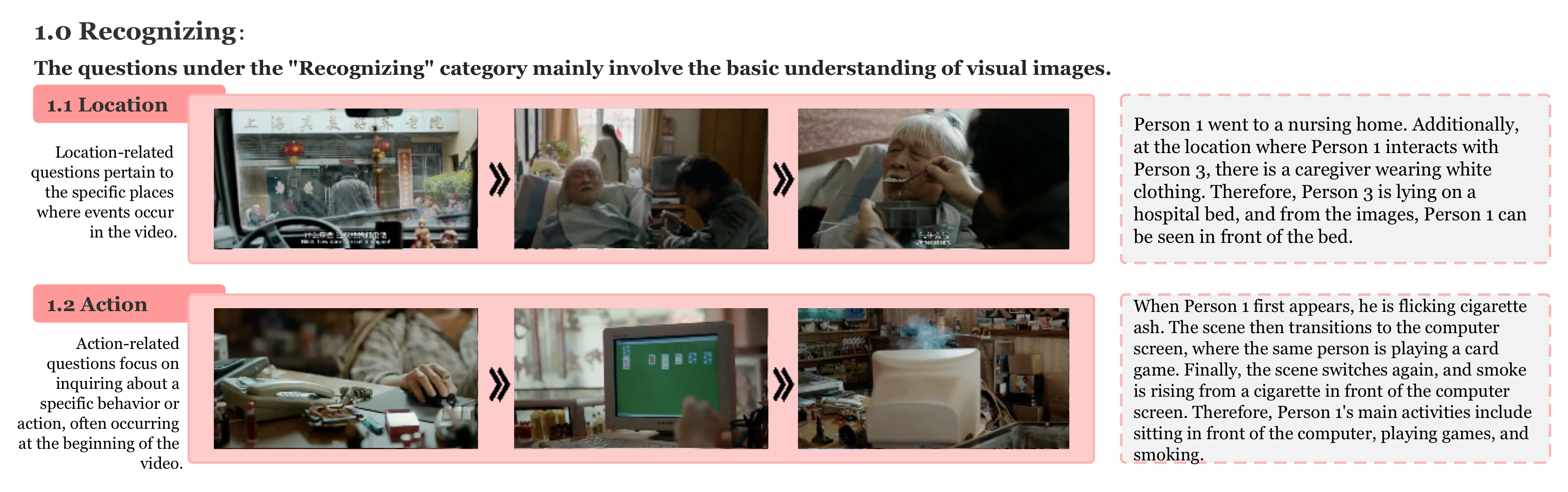}
    \caption{\textbf{CineClipQA}, the detailed description for the dataset.}
\label{fig:ccdes_1}
\end{figure*}
    \begin{figure*}
    \centering
    \includegraphics[width=\linewidth]{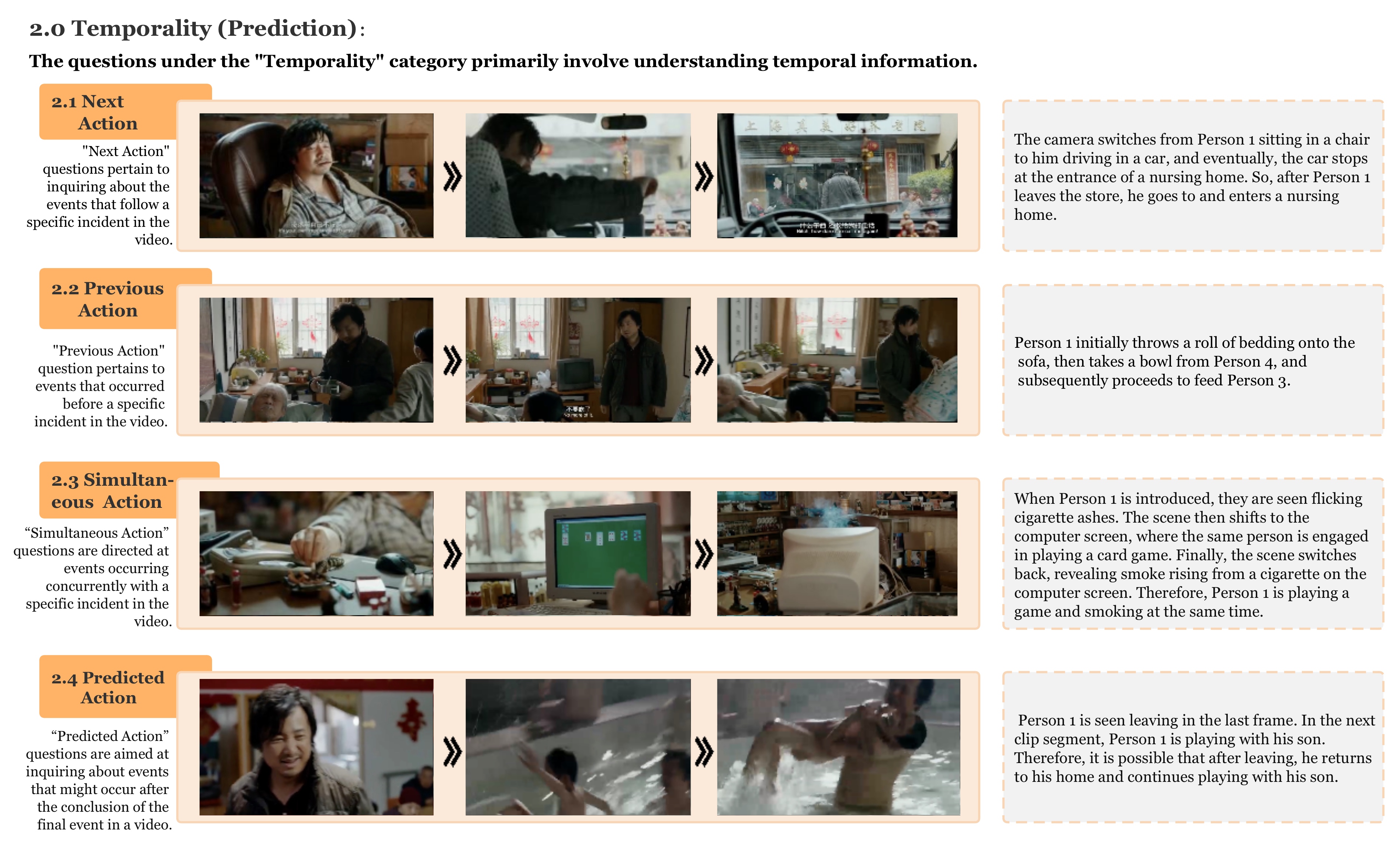}
    \caption{\textbf{CineClipQA}, the detailed description for the dataset.}
\label{fig:ccdes_2}
\end{figure*}
\begin{figure*}
    \centering
    \includegraphics[width=\linewidth]{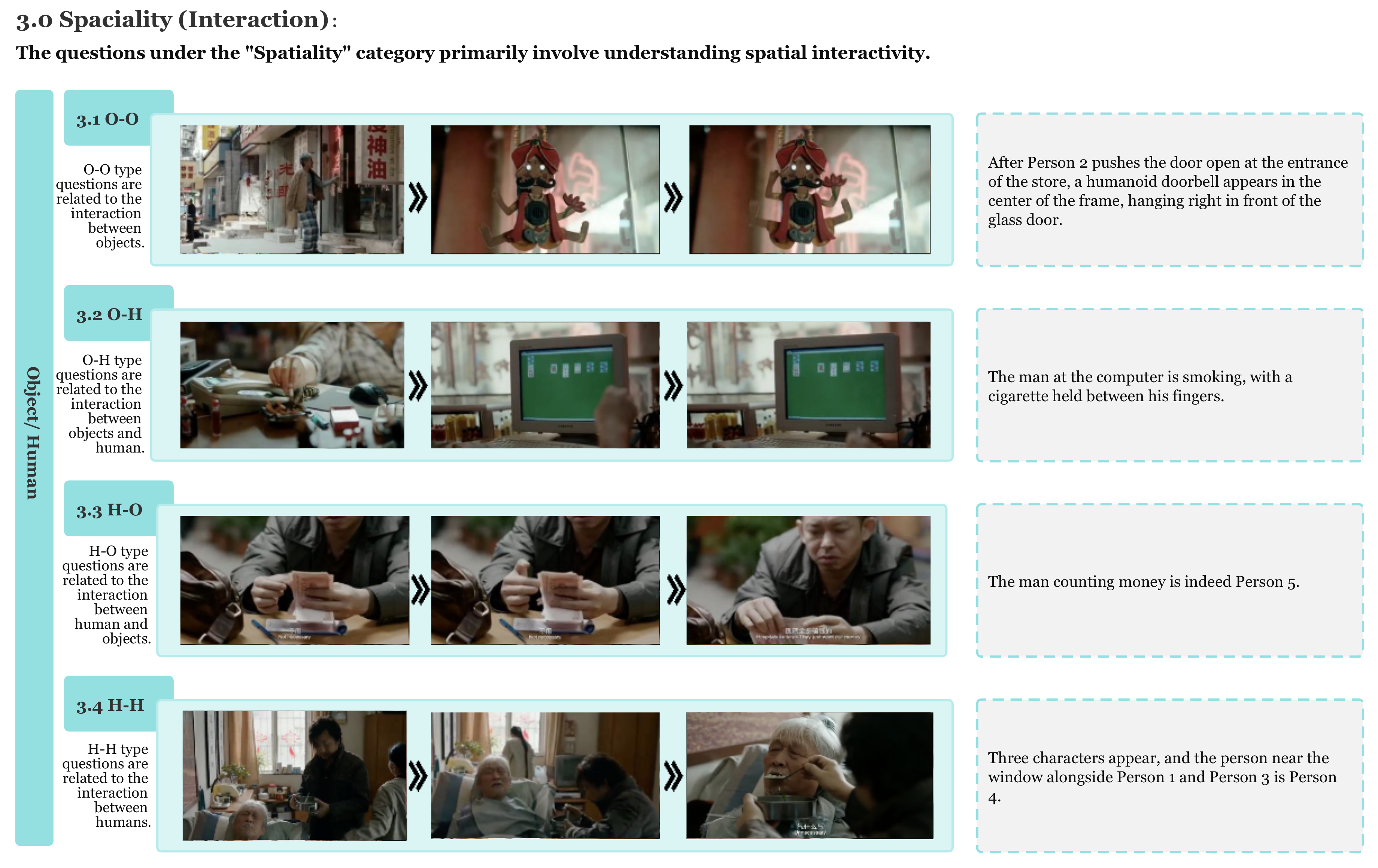}
    \caption{\textbf{CineClipQA}, the detailed description for the dataset.}
\label{fig:ccdes_3}
\end{figure*}
\begin{figure*}
    \centering
    \includegraphics[width=\linewidth]{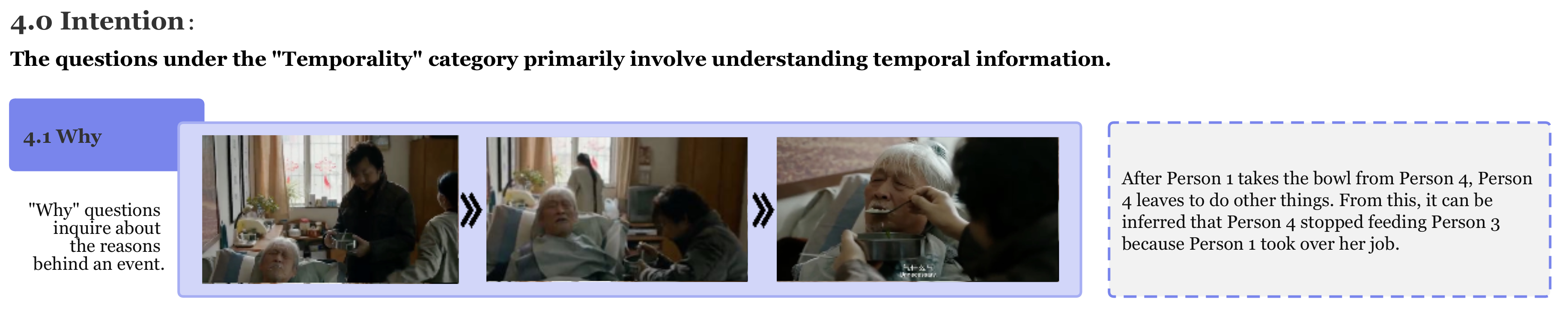}
    \caption{\textbf{CineClipQA}, the detailed description for the dataset.}
\label{fig:ccdes_4}
\end{figure*}
\begin{figure*}
    \centering
    \includegraphics[width=\linewidth]{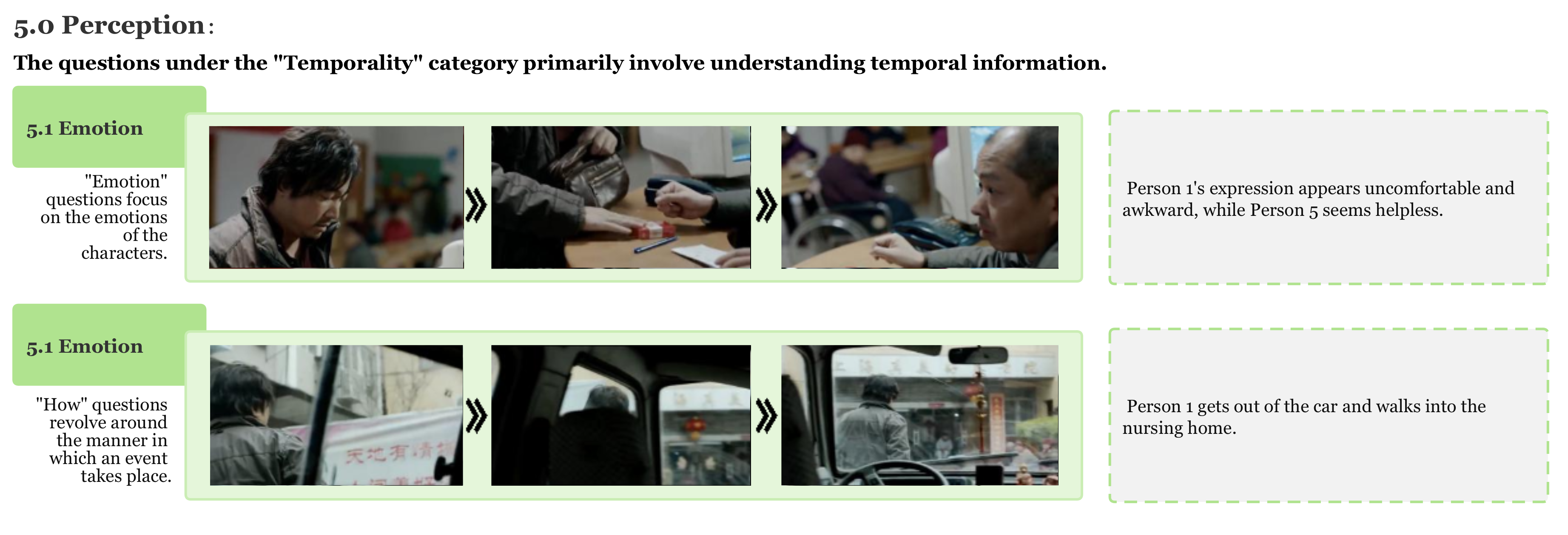}
    \caption{\textbf{CineClipQA}, the detailed description for the dataset.}
\label{fig:ccdes_5}
\end{figure*}
\noindent\textbf{Dataset Collection.}
In this paper, we introduce a new dataset named CineClipQA. The CineClipQA dataset encompasses a collection of 153 curated video clips, derived from five movies that span diverse genres and storytelling styles. Each clip, representing one or more distinct segment of the movie plot, is accompanied by a set of 16 tailored questions, thereby totaling 2,448 questions in various dimensions, as is presented in \cref{fig:ccdes}. The question consists of two parts: System Prompt and Questions. The System Prompt contains basic information about key characters in the current video clip and provides prompts for the initial actions of characters when necessary. The Questions are primarily divided into five categories: Recognizing, Temporality (Prediction), Spaciality (Interaction), Intention, and Perception. Specifically, Recognizing includes questions about Location and Action; Temporality encompasses questions about the next action, the previous action, simultaneous actions, and predicted actions; Spaciality involves questions about spatial information between Object and Human; Intention involves three similar types of questions about the purpose of actions; finally, Perception examines the recognition of emotions and inquiries about the "how" (approaches, manners...). At last, we provide a detailed explanation and corresponding case for all 16 types, shown in \cref{fig:ccdes_1}, \cref{fig:ccdes_2}, \cref{fig:ccdes_3}, \cref{fig:ccdes_4}, \cref{fig:ccdes_5}.

\noindent\textbf{Method Comparison.}
In our study, we evaluated the performance of several SOTA methods, including MovieChat, VideoLLAMA, Video-ChatGPT, and our own Vista-LLaMA, on the newly proposed CineClipQA dataset. This dataset, designed to assess comprehension of complex movie content, encompasses challenges in high content complexity, extensive scene variety, and prolonged temporal dimensions. The empirical results from our evaluations demonstrate that all methods, including our Vista-LLaMA, achieved state-of-the-art (SOTA) performance on the CineClipQA dataset. This outcome unequivocally confirms the superior ability of our approach in understanding videos with intricate content, wide-ranging scenes, and extended time frames.Notably, across all tested models, the highest accuracy was observed in the Intention category of the CineClipQA dataset. This suggests a particularly effective grasp of human behavioral reasoning, likely attributed to the rich prior knowledge embedded within these large language models (LLMs). The Intention category, by its nature, demands an in-depth analysis of purpose and motive behind actions depicted in the video clips, a task which seems to align well with the inherent strengths of current LLMs.Furthermore, this finding underscores the potential of LLMs in bridging the gap between mere visual recognition and deeper narrative understanding. The ability of these models to not only identify characters and actions but also infer underlying intentions is indicative of their advancing sophistication. It highlights a significant stride in the evolution of AI, where models are increasingly capable of nuanced interpretation akin to human-like understanding.

\section{Discussion}

\noindent\textbf{Advantages.}
In this study, we present two innovations, namely the EDVT-Attention and the sequential visual projector, aimed at enhancing video comprehension in LLMs. Our evaluation primarily focuses on zero-shot question answering benchmarks. The model used is built upon LLaVA~\cite{Liu2023VisualIT}, which is pre-trained and then fine-tuned with video instruction data. \model achieves a notable enhancement in the proposed innovations when tested on NExT-QA~\cite{Xiao2021NExTQANP} and MSRVTT-QA~\cite{Xu2016MSRVTTAL}. Additionally, we conduct several ablations to illustrate the effectiveness of our innovations. The outcomes demonstrate the significant potential of our approach to enhance video comprehension with LLMs.

\noindent\textbf{Limitations.}
There are also limitations in our work. For the VideoQA task, the evaluation process is assisted with GPT-3.5, which may result in some false judgments. GPT-4 might provide more accurate evaluations, but it comes at a higher cost since it is 20 times more expensive than GPT-3.5. Additionally, evaluating with GPT-4 requires the use of huge tokens, further increasing the expense. Furthermore, the evaluation speed is limited by query restrictions, and GPT-4 takes more time compared to training. We have evaluated a few cases using GPT-3.5, and the response has been reasonable and the accuracy has remained stable. When the same results are evaluated on NExT-QA three times, the variance is lower than 0.5 in the experiments.

Since this work only focuses on fine-tuning rather than pre-training, the full potential of EDVT-Attention may not be fully explored. EDVT-Attention can also be utilized for image-text related tasks. However, the impact of EDVT-Attention on pre-training, image-text related tasks, or other multi-modal tasks is not investigated in this manuscript. Additionally, the use of rotary positional embedding in some LLMs restricts the applicability of the current design. In this work, the rotary positional embedding is removed to ensure the same distance to visual tokens in decoder layers of LLMs. There may be alternative dynamic designs that can achieve this objective without eliminating the positional embedding. All these aspects are worth considering. Although the number of hallucination cases is reduced with our method, there are still instances where the model provides irrelevant replies. Further studies are necessary to address this issue. To enhance the current manuscript, our future work will focus on developing more general designs for practical cases.

\end{document}